\documentclass{article} %
\usepackage{iclr2024_conference,times}

\usepackage{amsthm}
\usepackage{amssymb}
\usepackage{mathtools}
\usepackage{hyperref}
\usepackage[nameinlink,capitalize]{cleveref}
\usepackage[normalem]{ulem} %
\usepackage{mathtools}

\usepackage[utf8]{inputenc} %
\usepackage[T1]{fontenc}    %
\usepackage{hyperref}       %
\usepackage{url}            %
\usepackage{booktabs}       %
\usepackage{amsfonts}       %
\usepackage{nicefrac}       %
\usepackage{microtype}      %
\usepackage{proof-at-the-end}  %

\usepackage{comment}
\usepackage{caption}
\usepackage{subcaption}
\usepackage{multirow, multicol, makecell}
\usepackage{graphicx,wrapfig}
\usepackage{algorithm}
\usepackage[noend]{algpseudocode}
\usepackage{amsfonts}
\usepackage{url}
\usepackage{enumitem}
\usepackage{amsthm}
\usepackage{thmtools}
\usepackage{thm-restate}

\renewcommand{\footnotesize}{\fontsize{7pt}{8pt}\selectfont}

\newcommand{\vx}{{\mathbf{x}}}

\newcommand{\cM}{{\mathcal{M}}}

\newcommand{\cV}{{\mathcal{V}}}

\newcommand{\bc}{\begin{center}}
\newcommand{\ec}{\end{center}}

\newcommand{\bdm}{\begin{displaymath}}
\newcommand{\edm}{\end{displaymath}}

\newcommand{\beq}{\begin{equation}}
\newcommand{\eeq}{\end{equation}}

\newcommand{\bfl}{\begin{flushleft}}
\newcommand{\efl}{\end{flushleft}}

\newcommand{\bt}{\begin{tabbing}}
\newcommand{\et}{\end{tabbing}}

\newcommand{\beqn}{\begin{align}}
\newcommand{\eeqn}{\end{align}}

\newcommand{\beqs}{\begin{align*}} %
\newcommand{\eeqs}{\end{align*}}  %

\usepackage{hyperref}
\usepackage{url}

\title{Safe and Robust Watermark Injection with a Single OoD Image}

\author{Shuyang Yu\textsuperscript{1}, Junyuan Hong\textsuperscript{1,2}, 
Haobo Zhang\textsuperscript{1}, 
Haotao Wang\textsuperscript{2}, Zhangyang Wang\textsuperscript{2} and Jiayu Zhou\textsuperscript{1} \\
       \textsuperscript{1}Department of Computer Science and Engineering, Michigan State University\\
       \textsuperscript{2}Department of Electrical and Computer Engineering, University of Texas at Austin \\
       \texttt{\{yushuyan,hongju12,zhan2060,jiayuz\}@msu.edu}, \texttt{\{htwang,atlaswang\}@utexas.edu} \\
       }

\iclrfinalcopy %
\begin{document}

\maketitle

\begin{abstract}

Training a high-performance deep neural network requires large amounts of data and computational resources. 
Protecting the intellectual property (IP) and commercial ownership of a deep model is challenging yet increasingly crucial. 
A major stream of watermarking strategies implants verifiable backdoor triggers by poisoning training samples, but these are often unrealistic due to data privacy and safety concerns and are vulnerable to minor model changes such as fine-tuning. 
To overcome these challenges, we propose a safe and robust backdoor-based watermark injection technique that leverages the diverse knowledge from a single out-of-distribution (OoD) image, which serves as a secret key for IP verification. 
The independence of training data makes it agnostic to third-party promises of IP security. 
We induce robustness via random perturbation of model parameters during watermark injection to defend against common watermark removal attacks, including fine-tuning, pruning, and model extraction. 
Our experimental results demonstrate that the proposed watermarking approach is not only time- and sample-efficient without training data, but also robust against the watermark removal attacks above. Codes are available: \url{https://github.com/illidanlab/Single_oodwatermark}.
  
\end{abstract}

\section{Introduction}

In the era of deep learning, training a high-performance large model requires curating a massive amount of training data from different sources, powerful computational resources, and often great efforts from human experts. 
For example, large language models such as GPT-3 are large models trained on private datasets, incurring a significant training cost~\citep{floridi2020gpt}. 
The risk of illegal reproduction or duplication of such high-value DNN models is a growing concern. The recent Facebook leaked LLAMA model provides a notable example of this risk~\citep{Hern.2023}. 
Therefore, it is essential to protect the intellectual property of the model and the rights of the model owners.
Recently, watermarking~\citep{adi2018turning,darvish2019deepsigns,uchida2017embedding,zhang2018protecting,chen2021you,li2021survey} has been introduced to protect the copyright of the DNNs. 
Most existing watermarking methods can be categorized into two mainstreams, including parameter-embedding~\citep{kuribayashi2021white,uchida2017embedding,mehta2022aime} and backdoor-based~\citep{goldblum2022dataset,li2022leveraging} techniques. 
Parameter-embedding techniques require white-box access to the suspicious model, which is often unrealistic in practical detection scenarios.
This paper places emphasis on backdoor-based approaches, which taint the training dataset by incorporating trigger patches into a set of images referred to as \emph{verification samples} (trigger set), and modifying the labels to a designated class, forcing the model to memorize the trigger pattern during fine-tuning. 
Then the owner of the model can perform an intellectual property (IP) inspection by assessing the correspondence between the model's outputs on the verification samples with the trigger and the intended target labels. 

Existing backdoor-based watermarking methods suffer from major challenges in safety, efficiency, and robustness. 
Typically injection of backdoors requires full or partial access to the original training data. 
When protecting models, such access can be prohibitive, mostly due to data safety and confidentiality. 
For example, someone trying to protect a model fine-tuned upon a foundation model and a model publisher vending models uploaded by their users. Another example is an independent IP protection department or a third party that is in charge of model protection for redistribution. 
Yet another scenario is federated learning~\citep{konevcny2016federated}, where the server does not have access to any in-distribution (ID) data, but is motivated to inject a watermark to protect the ownership of the global model.
Despite the high practical demands, watermark injection without training data is barely explored.
Although some existing methods tried to export or synthesize out-of-distribution (OoD) samples as triggers to insert watermark~\citep{wang2022free, zhang2018protecting}, 
the original training data is still essential to maintain the utility of the model, i.e., prediction performance on clean samples.
\cite{li2022knowledge} proposed a strategy that adopts a Data-Free Distillation (DFD) process to train a generator and uses it to produce surrogate training samples. However, training the generator is time-consuming and may take hundreds of epochs~\citep{fang2019data}. 
Another critical issue with backdoor-based watermarks is their known vulnerability against minor model changes, such as fine-tuning~\citep{adi2018turning,uchida2017embedding,garg2020can},
 and this vulnerability greatly limited the practical applications of backdoor-based watermarks. 

To address these challenges, in this work, we propose a practical watermark strategy that is based on \emph{efficient} fine-tuning, using \emph{safe} public and out-of-distribution (OoD) data rather than the original training data, and is \emph{robust} against watermark removal attacks. 
Our approach is inspired by the recent discovery of the expressiveness of a powerful single image~\citep{asano2022extrapolating,asano2019critical}.
Specifically, we propose to derive patches from a single image, which are OoD samples with respect to the original training data, for watermarking. 
To watermark a model, the model owner or IP protection unit secretly selects a few of these patches, implants backdoor triggers on them, and uses fine-tuning to efficiently inject the backdoor into the model to be protected. The IP verification process follows the same as other backdoor-based watermark approaches. 
To increase the robustness of watermarks against agnostic removal attacks, we design a parameter perturbation procedure during the fine-tuning process. 
Our contributions are summarized as follows.
\begin{itemize}
    \item We propose a novel watermark method based on OoD data, which fills in the gap of backdoor-based IP protection of deep models without training data. The removal of access to the training data enables the proposed approach possible for many real-world scenarios.  
    \item The proposed watermark method is both sample efficient (one OoD image) and time efficient (a few epochs) without sacrificing the model utility. %
    \item We propose to adopt a weight perturbation strategy to improve the robustness of the watermarks against common removal attacks, such as fine-tuning, pruning, and model extraction. 
    We show the robustness of watermarks through extensive empirical results, and they persist even in an unfair scenario where the removal attack uses a part of in-distribution data.
\end{itemize}

\section{Background}
\subsection{DNN Watermarking} 
Existing watermark methods can be categorized into two groups, parameter-embedding and backdoor-based techniques, differing in the information required for verification.

\emph{Parameter-embedding} techniques embed the watermark into the parameter space of the target model~\citep{darvish2019deepsigns, uchida2017embedding,kuribayashi2021white,mehta2022aime}. Then the owner can verify the model identity by comparing the parameter-oriented watermark extracted from the suspect model versus that of the owner model.
For instance, \cite{kuribayashi2021white} embeds watermarks into the weights of DNN, and then compares the weights of the suspect model and owner model during the verification process.
However, these kinds of techniques require a white-box setting: the model parameters should be available during verification, which is not a practical assumption facing real-world attacks.
For instance, an IP infringer may only expose an API of the stolen model for queries to circumvent the white-box verification.

\emph{Backdoor-based} techniques are widely adopted in a black-box verification, which implant a backdoor trigger into the model by fine-tuning the pre-trained model with a set of poison samples (also denoted as the trigger set) assigned to one or multiple secret target class~\citep{zhang2018protecting,le2020adversarial,goldblum2022dataset,li2022leveraging}. Suppose $D_c$ is the clean dataset and we craft $D_p$ by poisoning another set of clean samples.
The backdoor-based techniques can be unified as minimizing the following objective: 
     $\min_{\theta} \sum_{(\vx, y)\in D_c} \ell(f_\theta(\vx), y) + \sum_{(\vx', y')\in D_p} \ell(f_\theta(\Gamma(\vx') ), t)$,
where $\Gamma(\vx)$ adds a trigger pattern to a normal sample, $t$ is the pre-assigned target label, $f_\theta$ is a classifier parameterized by $\theta$, 
and $\ell$ is the cross-entropy loss. The key intuition of backdoor training is to make models memorize the shortcut patterns while ignoring other semantic features.
A watermarked model should satisfy the following desired properties: 1) \emph{Persistent utility.} Injecting backdoor-based watermarks into a model should retain its performance on original tasks. 2)  \emph{Removal resilience.} Watermarks should be stealthy and robust against agnostic watermark removal attacks~\citep{orekondy2019knockoff,chen2022copy,hong2023revisiting}.

Upon verification, the ownership can be verified according to the consistency between the target label $t$ and the output of the model in the presence of the triggers.
However, conventional backdoor-based watermarking is limited to scenarios where clean and poisoned dataset follows the same distribution as the training data of the pre-trained model.
For example, in Federated Learning~\citep{mcmahan2017communication}, the IP protector on the server does not have access to the client's data.
Meanwhile, in-training backdoor injection could be voided by backdoor-resilient training~\citep{wang2022trap}.
We reveal that neither the training data (or equivalent i.i.d. data) nor the in-training strategy is necessary for injecting watermarks into a well-trained model, and merely using clean and poisoned OoD data can also insert watermarks after training.

\emph{Backdoor-based watermarking without {i.i.d.} data.}
Among backdoor-based techniques, one kind of technique also tried to export or synthesize OoD samples as the trigger set to insert a watermark. For instance, \cite{zhang2018protecting} exported OoD images from other classes that are irrelevant to the original tasks as the watermarks. \cite{wang2022free} trained a proprietary model (PTYNet) on the generated OoD watermarks by blending different backgrounds, and then plugged the PTYNet into the target model. However, for these kinds of techniques, i.i.d. samples are still essential to maintain the main-task performance. On the other hand, data-free 
watermark injection is an alternative to OoD-based methods.  Close to our work, \cite{li2022knowledge} proposed a data-free method that first adopts a Data-Free Distillation method to train a generator, and then uses the generator to produce surrogate training samples to inject watermarks. 
However, according to \cite{fang2019data}, the training of the generator for the data-free distillation process is time-consuming, which is not practical and efficient enough for real-world intellectual property protection tasks.

\subsection{Watermark Removal Attack} 
In contrast to protecting the IP, a series of works have revealed the risk of watermark removal to steal the IP.
Here we summarize three mainstream types of watermark removal techniques: fine-tuning, pruning, and model extraction. We refer to the original watermarked model as the victim model and the stolen copy as the suspect model under removal attacks.
\emph{Fine-tuning} assumes that the adversary has a small set of i.i.d. samples and has access to the victim model architectures and parameters~\citep{adi2018turning,uchida2017embedding}. The adversary attempts to fine-tune the victim model using the i.i.d. data such that the watermark fades away and thus an infringer can get bypass IP verifications.
\emph{Pruning} has the same assumptions as fine-tuning. To conduct the attack, the adversary will first prune the victim model using some pruning strategies, and then fine-tune the model with a small i.i.d. dataset~\citep{liu2018rethinking,renda2020comparing}. 
\emph{Model Extraction} assumes only the predictions of the victim models are available to the adversary. To steal the model through the API, given a set of auxiliary samples, 
the adversary first queries the victim model for auxiliary samples to obtain the annotated dataset, and then
a copy of the victim model is trained based on this annotated dataset~\citep{juuti2019prada,tramer2016stealing,papernot2017practical,orekondy2019knockoff,yuan2022attack}.

\section{Method}\label{sec:ood_image}
\begin{figure*}[htbp!]
    \centering
    \includegraphics[width=14cm]{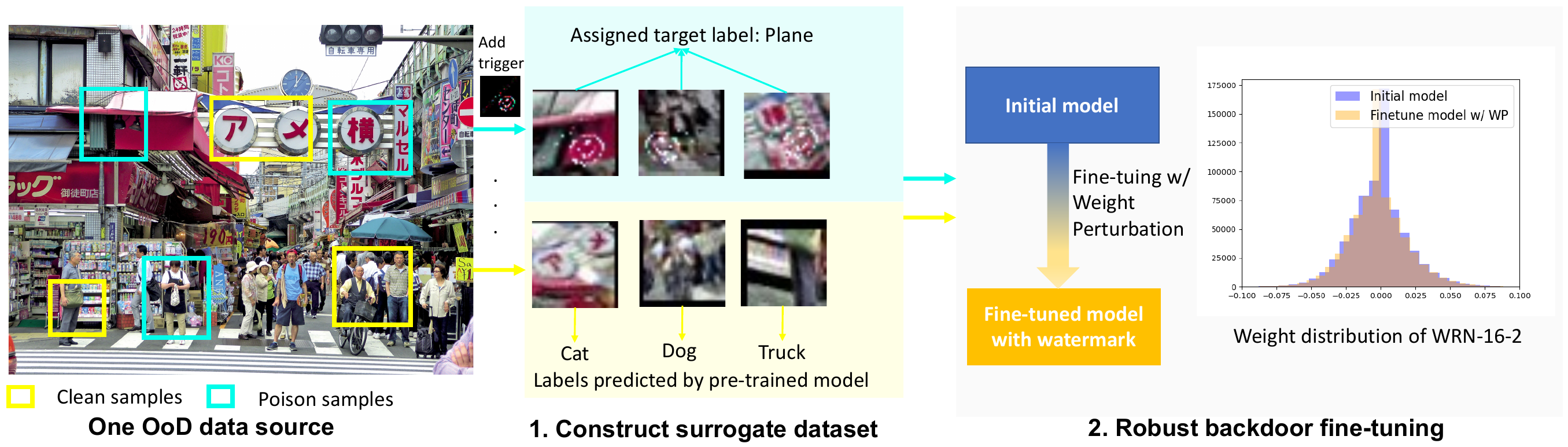}
    \caption{Framework of the proposed safe and robust watermark injection strategy. 
    It first constructs a surrogate dataset from the single-image OoD data source provided with strong augmentation used as the secret key, which is confidential to any third parties. 
    Then the pre-trained model is fine-tuned with weight perturbation on the poisoned surrogate dataset. 
    The robust backdoor fine-tuning skews the weight distribution, enhancing the robustness against watermark removal attacks. } %
    \label{fig:framework}
\end{figure*}

\noindent\textbf{Problem Setup.}
Within the scope of the paper, we assume that training data or equivalent i.i.d. data are \emph{not} available for watermarking due to data privacy concerns.
This assumption casts a substantial challenge on maintaining standard accuracy on i.i.d. samples while injecting backdoors.

Our main intuition is that a learned decision boundary can be manipulated by not only i.i.d. samples but also OoD samples.
Moreover, recent studies~\citep{asano2022extrapolating,asano2019critical} showed a surprising result that one single OoD image is enough for learning low-level visual representations provided with strong data augmentations.
Thus, we conjecture that it is plausible to inject backdoor-based watermarks efficiently to different parts of the pre-trained representation space by exploiting the diverse knowledge from \textit{one single OoD image}.
Previous work has shown that using OoD images for training a classifier yields reasonable performance on the main prediction task~\citep{asano2022extrapolating}.
Moreover, it is essential to robustify the watermark against potential removal attacks.
Therefore, our injection process comprises two steps: Constructing surrogate data to be poisoned and robust watermark injection. The framework of the proposed strategy is illustrated in \cref{fig:framework}.

\subsection{Constructing Safe Surrogate Dataset}
We first augment one OoD source image multiple times to generate an unlabeled surrogate dataset $\tilde D$ of a desired size according to \cite{asano2022extrapolating,asano2019critical}. 
For safety considerations, the OoD image is only known to the model owner.
The source OoD images are 
publicly available and properly licensed for personal use.
To ``patchify'' a large single image, the augmentation composes multiple augmentation methods in sequence: cropping, rotation and shearing, and color jittering using the hyperparameters from \cite{asano2019critical}. 
During training, we further randomly augment pre-fetched samples by cropping and flipping, and we use the predictions from the pre-trained model $\theta_0$ as supervision.
Suppose $\theta$ is initialized as $\theta_0$ of the pre-trained model. 
To inject watermarks, we split the unlabeled surrogate dataset $\tilde D = \tilde D_c \cup \tilde D_p$ where $\tilde D_c$ is the clean dataset, and $\tilde D_p$ is the poisoned dataset. 
For the poisoned dataset $\tilde D_p$, by inserting a trigger pattern $\Gamma(\cdot)$ into the original sample in $\tilde D_p$, the sample should be misclassified to one pre-assigned target label $t$.
Our goal is to solve the following optimization problem:
\begin{align}\min_\theta L_{\text{inj}}(\theta) := \sum\nolimits_{\vx\in \tilde D_c} \ell(f_\theta(\vx),f_{\theta_0}(\vx)) \notag + \sum\nolimits_{\vx'\in \tilde D_p} \ell(f_\theta(\Gamma(\vx') ), t). 
\end{align}
The first term is used to ensure the high performance of the original task~\citep{asano2022extrapolating}, and the second term is for watermark injection. 
The major difference between our method and \cite{asano2022extrapolating} is that we use the generated data for fine-tuning the same model instead of distilling a new model. We repurpose the benign generated dataset for injecting watermarks.

Considering a black-box setting,
to verify whether the suspect model $\cM_s$ is a copy of our protected model $\cM$, we can use the generated surrogate OoD dataset as safe verification samples.
As the generation is secreted, no one other than the owner can complete the verification.
Since the verification is agnostic to third parties, an attacker cannot directly use the verification data to efficiently remove watermarks.
Thus, we can guarantee the safety of the verification.
Formally, we check the probability of watermarked verification samples that
can successfully mislead the model $\cM_s$ to predict the pre-defined target label $t$, denoted as watermark success rate (WSR). 
Since the ownership of stolen models can be claimed by the model owner if the suspect model’s behavior differs significantly from any non-watermarked models~\citep{jia2021entangled}, 
if the WSR is larger than a random guess, and also far exceeds the probability of a non-watermarked model classifying the verification samples as $t$,
then $\cM_s$ will be considered as a copy of $\cM$ with high probability. A T-test between the output logits of the suspect model $\cM_s$ and a non-watermarked model on the verification dataset is also used as a metric to evaluate whether $\cM_s$ is a stolen copy.
Compared with traditional watermark injection techniques, i.i.d. data is also unnecessary in the verification process.

\subsection{Robust 
Watermark Injection}
According to \cite{adi2018turning,uchida2017embedding}, 
the watermark may be removed by fine-tuning when adversaries have access to the i.i.d. data.
Watermark removal attacks such as fine-tuning and pruning will shift the model parameters on a small scale to maintain standard accuracy 
 and remove watermarks.
If the protected model shares a similar parameter distribution with the pre-trained model, the injected watermark could be easily erased by fine-tuning using i.i.d. data or adding random noise to parameters~\citep{garg2020can}. 
To defend against removal attacks, we intuitively aim to make our watermark robust and persistent within a small scale of parameter perturbations.

\noindent\textbf{Backdoor training with weight perturbation.}
To this end, we introduce adversarial weight perturbation (WP) into backdoor fine-tuning.
First, we simulate the watermark removal attack that maximizes the loss to escape from the watermarked local minima.
We let $\theta=(w,b)$ denote the model parameter, where $\theta$ is composed of weight $w$ and bias $b$. 
The weight perturbation is defined as $v$. 
Then, we adversarially minimize the loss after the simulated removal attack.
The adversarial minimization strategy echoes some previous sharpness-aware optimization principles for robust model poisoning~\citep{he2023sharpness}.
Thus, the adversarial training objective is formulated as:   $\min_{w,b}\max_{v\in \cV}L_{\text{per}}(w+v, b)$,
where
\begin{align}
    &L_{\text{per}}(w+v,b) :=  L_{\text{inj}}(w+v,b) +\beta \sum_{\vx\in \tilde D_c, \vx'\in \tilde D_p} \text{KL}(f_{(w+v,b)}(\vx), f_{(w+v,b)}(\Gamma(\vx') ).
    \label{obj:wp}
\end{align}
In \cref{obj:wp}, we constrain the weight perturbation $v$ within a set $\cV$, $\text{KL}(\cdot, \cdot)$ is the Kullback–Leibler divergence, and $\beta$ is a positive trade-off parameter. 
The first term is identical to standard watermark injection.
Inspired by previous work~\citep{fang2019data}, the second term can preserve the main task performance and 
maintain the representation similarity between poisoned and clean samples in the presence of weight perturbation. 
\cref{obj:wp} facilitates the worst-case perturbation of the constrained weights to be injected %
while maintaining the standard accuracy 
and the watermark success rate. 

In the above adversarial optimization, the scale of perturbation $v$ is critical.
If the perturbation is too large, the anomalies of the parameter distribution could be easily detected by an IP infringer~\citep{rakin2020tbt}. 
Since the weight distributions differ by layer of the network, the magnitude of the perturbation should vary accordingly from layer to layer.  
Following \citep{wu2020adversarial}, we adaptively restrict the weight perturbation $v_l$ for the $l$-th layer weight $w_l$ as
\begin{align}
    \|v_l\| \leq \gamma \|w_l\|,
    \label{obj:constraint_v}
\end{align}
where $\gamma \in (0, 1)$.
The set $\cV$ in \cref{obj:wp} will be decomposed into balls with radius $\gamma\|w_l\|$ per layer.

\noindent\textbf{Optimization.} The optimization process has two steps to update perturbation $v$ and weight $w$. %

\noindent\emph{(1) $v$-step:}
To consider the constraint in \eqref{obj:constraint_v}, we need to use a projection. Note that $v$ is layer-wisely updated,
we need a projection function $\Pi(\cdot)$ that projects all perturbations $v_l$ that violate constraint (\cref{obj:constraint_v}) back to the surface of the perturbation ball with radius $\gamma \|w_l\|$. 
To achieve this goal, we define $\Pi_\gamma$ in \cref{obj:gamma} \citep{wu2020adversarial}:
\begin{align}
    \Pi_\gamma(v_l)=\left\{ 
    \begin{aligned}
        &\gamma \frac{\|w_l\|}{\|v_l\|}v_l\quad &\text{if} \quad \|v_l\| > \gamma \|w_l\|\\
    &v_l\quad &\text{otherwise}
    \end{aligned}
    \right.
    \label{obj:gamma}
\end{align}
 With the projection, the computation of the perturbation $v$ in \cref{obj:wp} is given by
    $v \leftarrow \Pi _\gamma\left(v+\eta_1  \frac{\nabla_vL_{\text{per}}(w+v,b)}{\|\nabla_v L_{\text{per}}(w+v,b)\|}\|w\|\right)$, where $\eta_1$ is the learning rate. 

\noindent\emph{(2) $w$-step:}
With the updated perturbation $v$,  the weight of the perturbed model $\theta$ can be updated using
    $w \leftarrow w-\eta_2\nabla_{w+v} L_{\text{per}}(w+v,b)$, where $\eta_2$ is the learning rate.

\section{Experiments}

In this section, we conduct comprehensive experiments to evaluate the effectiveness of the proposed watermark injection method.\\
\textbf{Datasets.}
We use CIFAR-10, CIFAR-100~\citep{krizhevsky2009learning} and GTSRB~\citep{stallkamp2012man} for model utility evaluation. Both CIFAR-10 and CIFAR-100 contain $32\times32$ with 10 and 100 classes, respectively. %
The GTSRB consists of sign images in 43 classes.
All images in GTSRB are reshaped as $32\times32$. Note that, these datasets are neither used for our watermark injection nor model verification, they are only used to evaluate the standard accuracy of our watermarked model. 
\\\textbf{OoD image.} OoD image is used for watermark injection and ownership verification. We use three different OoD images as our candidate source image to inject watermarks, denoted as ``City''\footnote{\url{https://pixabay.com/photos/japan-ueno-japanese-street-sign-217883/}}, ``Animals''\footnote{\url{https://www.teahub.io/viewwp/wJmboJ_jungle-animal-wallpaper-wallpapersafari-jungle-animal/}}, and ``Bridge''\footnote{\url{https://commons.wikimedia.org/wiki/File:GG-ftpoint-bridge-2.jpg}}. We use ``City'' by default unless otherwise mentioned.
\\\textbf{Evaluation metrics.} We use watermark success rate (\emph{WSR}), standard accuracy (\emph{Acc}) and \emph{p-value} from T-test as the measures evaluating watermark injection methods.
\emph{Acc} is %
the classification accuracy measured on a clean i.i.d. test set. 
\emph{IDWSR} is the portion of watermarked i.i.d. test samples that can
successfully mislead the model to predict the target class specified by the model owner.
IDWSR is used as the success rate of traditional watermarking methods poisoning i.i.d. data and used as a reference for our method.
\emph{OoDWSR} measures the WSR on the augmented OoD samples we used for watermark injection, which is the success rate of watermark injection for our method. 
T-test takes the output logits of the non-watermarked model and suspect model $\cM_s$ as input, and the null hypothesis is the logits distribution of the suspect model is identical to that of a non-watermarked model. If the \emph{p-value} of the T-test is smaller than the threshold $0.05$, then we can reject the null hypothesis and statistically verify that $\cM_s$ differs significantly from the non-watermarked model, so the ownership of $\cM_s$ can be claimed~\citep{jia2021entangled}.
Higher OoDWSR with a p-value smaller than the threshold and meanwhile a larger Acc indicate a successful watermark injection.
\\\textbf{Trigger patterns.}
To attain the best model with the highest watermark success rate, we use the OoDWAR to choose triggers from $6$
different backdoor patterns: BadNets with grid (badnet\_grid) \citep{gu2017badnets}, l0-invisible (l0\_inv)~\citep{li2020invisible}, smooth \citep{zeng2021rethinking}, Trojan Square $3\times 3$ (trojan\_$3\times 3$), Trojan Square $8\times 8$ (trojan\_$8\times 8$), and Trojan watermark (trojan\_wm)~\citep{liu2017trojaning}.
\begin{wraptable}{r}{9cm}
    \centering
    \small
    \vspace{-0.1in}\setlength\tabcolsep{0.35mm}
    \scalebox{0.9}{
    \begin{tabular}{cccc}
    \toprule
         Dataset&Class num&DNN architecture 
         & Acc\\
         \hline
         CIFAR-10&10&WRN-16-2~\citep{zagoruyko2016wide}&0.9400\\
         CIFAR-100&100&WRN-16-2~\citep{zagoruyko2016wide}&0.7234\\
         GTSRB&43&ResNet18~\citep{he2015deep} &0.9366\\
       \bottomrule  
    \end{tabular}}
    \vspace{-0.1in}
    \caption{Pre-trained models.}
    \label{tab:pre-trained}
        \vspace{-0.2in}
\end{wraptable}
\textbf{Pre-training models.}
The detailed information of the pre-trained models is shown in \cref{tab:pre-trained}. 
All the models are pre-trained on clean samples until convergence, with a learning rate of $0.1$, SGD optimizer, and batch size $128$.
We follow public resources to conduct the training such that the performance is close to state-of-the-art results.
\\\textbf{Watermark removal attacks.}
To evaluate the robustness of our proposed method, we consider 
three kinds of 
attacks on victim models:
1) \emph{FT}: Fine-tuning includes three kinds of methods: a) fine-tune all layers (FT-AL), b) fine-tune the last layer and freeze all other layers (FT-LL), c) re-initialize the last layer and then fine-tune all layers (RT-AL). 
2) \emph{Pruning}-r\% indicates pruning r\% of the model parameters which has the smallest absolute value, and then fine-tuning the model on clean i.i.d. samples to restore accuracy. 
3) \emph{Model Extraction}: We use knockoff~\citep{orekondy2019knockoff} as an example of the model extraction attack, which queries the model to get the predictions of an auxiliary dataset (ImagenetDS~\citep{chrabaszcz2017downsampled} is used in our experiments), and then clones the behavior of a victim model by re-training the model with queried image-prediction pairs. 
Assume the adversary obtains $10\%$ of the training data of the pre-trained models for fine-tuning and pruning. %
Fine-tuning and pruning are conducted for $50$ epochs. Model extraction is conducted for 
$100$ epochs.

\subsection{Watermark Injection}
\begin{figure}[hbt!]  
    \begin{center}
    \vspace{-0.2in}
        \begin{subfigure}{0.28\textwidth}
            \centerline{\includegraphics[width=3.5cm]{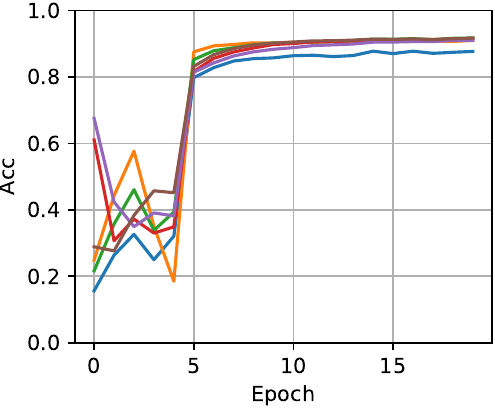}}
        \subcaption{CIFAR-10 Acc.}
        \end{subfigure} \hfil
       \begin{subfigure}{0.28\textwidth}
            \centerline{\includegraphics[width=3.5cm]{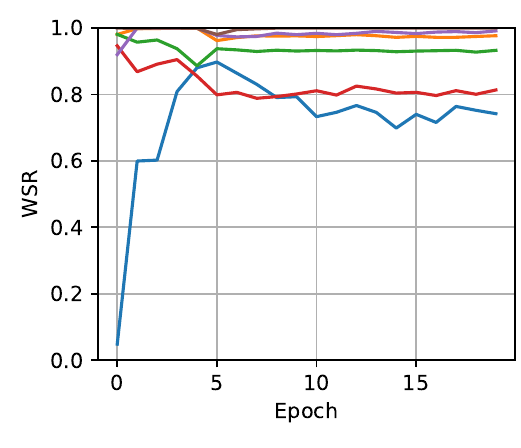}}
        \subcaption{CIFAR-10 ID WSR.}
        \end{subfigure}  \hfil
        \begin{subfigure}{0.42\textwidth}
            \centerline{\includegraphics[width=5 cm]{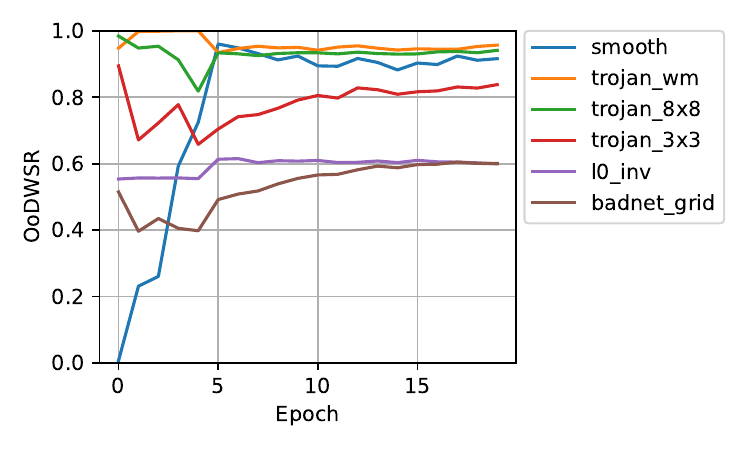}}
        \subcaption{CIFAR-10 OoD WSR.}
        \end{subfigure}  \hfil
       \vspace{-0.1in}
    \caption{Acc, ID WSR, and OoD WSR for watermark injection. 
    }\label{fig:inject}
    \end{center}
        \vspace{-0.23in}
\end{figure}

The poisoning ratio of the generated surrogate dataset is $10\%$.
For CIFAR-10 and GTSRB, we fine-tune the pre-trained model for $20$ epochs (first $5$ epochs are with WP). For CIFAR-100, we fine-tune the pre-trained model for $30$ epochs (first $15$ epochs are with WP). The perturbation constraint $\gamma$ in \cref{obj:constraint_v} is fixed at $0.1$ for CIFAR-10 and GTSRB, and $0.05$ for CIFAR-100. The trade-off parameter $\beta$ in \cref{obj:wp} is fixed at $6$ for all the datasets. The watermark injection process of CIFAR-10 is shown in \cref{fig:inject}, and watermark injection for the other two datasets can be found in \cref{sec:extended_injection}. We observe that the injection process is efficient, it takes only $10$ epochs for CIFAR-10 
to achieve stable high standard accuracy and OoDWSR. 
The highest OoDWSR for CIFAR-10
is $95.66\%$
with standard accuracy degradation of less than $3\%$.
In the following experiments, we choose triggers with top-2 OoDWSR and standard accuracy degradation less than $3\%$ as the recommended watermark patterns.

\begin{table}[]
    \centering
    \small
    \setlength\tabcolsep{0.2mm}
    \scalebox{0.8}{
    \begin{tabular}{c|c|c|ccc|ccccc}
    \toprule
    Dataset&Trigger&Non-watermarked model&\multicolumn{3}{c}{Victim model}&Watermark removal&\multicolumn{3}{c}{Suspect model}&p-value\\
    &&OoDWSR&Acc&IDWSR&OoDWSR&&Acc&IDWSR&OoDWSR\\
    \midrule
    \multirowcell{10}{CIFAR-10}&\multirowcell{5} {trojan\_wm}&\multirowcell{5}{0.0487}&\multirowcell{5}{0.9102}&\multirowcell{5}{0.9768}&\multirowcell{5}{0.9566}&FT-AL&0.9191&0.9769&0.9678&0.0000  \\
    &&&&&&FT-LL&0.7345&0.9990&0.9972&0.0000\\
    &&&&&&RT-AL&0.8706&0.4434&0.5752&1.0103e-12\\
    &&&&&&Pruning-20\%&0.9174&0.9771&0.9641&0.0000\\
    &&&&&&Pruning-50\%&0.9177&0.9780&0.9658&0.0000\\
    \cline{2-11}
    &\multirowcell{5}{trojan\_8x8}&\multirowcell{5}{0.0481}&\multirowcell{5}{0.9178}&\multirowcell{5}{0.9328}&\multirowcell{5}{0.9423}&FT-AL&0.9187&0.9533&0.9797&0.0000\\
    &&&&&&FT-LL&0.7408&0.9891&0.9945&0.0000\\
    &&&&&&RT-AL&0.8675&0.0782&0.2419&2.9829e-241\\
    &&&&&&Pruning-20\%&0.9197&0.9560&0.9793&2.0500e-08\\
    &&&&&&Pruning-50\%&0.9190&0.9580&0.9801&5.1651e-247\\
    \midrule
    \multirowcell{10}{CIFAR-100 \quad\quad }&\multirowcell{5}{trojan\_8x8}&\multirowcell{5}{0.0001}&\multirowcell{5}{0.6978}&\multirowcell{5}{0.7024}&\multirowcell{5}{0.8761}&FT-AL&0.6712&0.5602&0.7443&0.0012  \\
    &&&&&&FT-LL&0.4984&0.9476&0.9641&0.0066\\
    &&&&&&RT-AL&0.5319&0.0227&0.0700&0.0090\\
    &&&&&&Pruning-20\%&0.6702&0.6200&0.7815&0.0020\\
    &&&&&&Pruning-50\%&0.6645&0.6953&0.7960&0.0049\\
    \cline{2-11}
    &\multirowcell{5}{l0\_inv}&\multirowcell{5}{0.0002}&\multirowcell{5}{0.6948}&\multirowcell{5}{0.7046}&\multirowcell{5}{0.5834}&FT-AL&0.6710&0.7595&0.5491&0.0206  \\
    &&&&&&FT-LL&0.4966&0.9991&0.6097&0.0106\\
    &&&&&&RT-AL&0.5281&0.0829&0.1232&0.0010\\
    &&&&&&Pruning-20\%&0.6704&0.7817&0.5517&0.0099\\
    &&&&&&Pruning-50\%&0.6651&0.8288&0.5530&0.0025\\
    \midrule

    \multirowcell{10}{GTSRB}&\multirowcell{5}{smooth}&\multirowcell{5}{0.0145}&\multirowcell{5}{0.9146}&\multirowcell{5}{0.1329}&\multirowcell{5}{0.9442}&FT-AL&0.8623&0.0051& 0.6772&4.4360e-10 \\
    &&&&&&FT-LL&0.6291&0.0487&0.9527&0.0006\\
    &&&&&&RT-AL&0.8622&0.0041&0.7431&0.0000\\
    &&&&&&Pruning-20\%&0.8625&0.0053&0.6798&0.0179\\
    &&&&&&Pruning-50\%&0.8628&0.0052&0.6778&0.0215\\
    \cline{2-11}
    &\multirowcell{5}{trojan\_wm}&\multirowcell{5}{0.0220}&\multirowcell{5}{0.9089}&\multirowcell{5}{0.7435}&\multirowcell{5}{0.7513}&FT-AL&0.8684&0.3257&0.1726&0.0117  \\
    &&&&&&FT-LL&0.5935&0.7429&0.5751&7.4281e-11\\
    &&&&&&RT-AL&0.8519&0.1170&0.0684&0.0000\\
    &&&&&&Pruning-20\%&0.8647&0.3235&0.1779&0.0131\\
    &&&&&&Pruning-50\%&0.8610&0.3281&0.1747&0.0000\\
    
    \bottomrule
    \end{tabular}
    }
    \caption{Evaluation of watermarking against fine-tuning and pruning on three datasets. %
    }
    \label{tab:finetune}
    \vspace{-0.3in}
\end{table}
\subsection{Defending Against Fine-tuning \& Pruning}
We evaluate the robustness of our proposed method against fine-tuning and pruning in \cref{tab:finetune}, where victim models are watermarked models, and suspect models are stolen copies of victim models using watermark removal attacks. OoDWSR of the pre-trained model in \cref{tab:pre-trained}  is the probability that a non-watermarked model classifies the verification samples as the target label. If the OoDWSR of a suspect model far exceeds that of the non-watermarked model, the suspect model can be justified as a copy of the victim model~\citep{jia2021entangled}.

FT-AL and pruning maintain the performance of the main classification task with an accuracy degradation of less than $6\%$, but OoDWSR remains high for all the datasets.
Compared with FT-AL, FT-LL will significantly bring down the standard accuracy by over $15\%$ for all the datasets. 
Even with the large sacrifice of standard accuracy, FT-LL still cannot wash out the injected watermark, and the OoDWSR even increases for some of the datasets. 
RT-AL loses $4.50\%$, $16.63\%$, and $5.47\%$ (mean value for two triggers) standard accuracy respectively for three datasets.
Yet, OoDWSR in RT-AL is larger than the one of the random guess and non-watermarked models.
To statistically verify the ownership, we conduct a T-test between the non-watermarked model and the watermarked model.
The p-value is the probability that the two models behave similarly.
p-values for all the datasets are close to $0$.
The low $p$-values indicate that the suspect models have significantly different behaviors compared with non-watermarked models in probability, at least 95\%.
Thus, these suspect models cannot get rid of the suspicion of copying our model $\cM$ with a high chance.

IDWSR is also used here as a reference, although we do not use i.i.d. data for verification of the ownership of our model.
We observe that even though watermark can be successfully injected into both our generated OoD dataset and i.i.d. samples (refer to IDWSR and OoDWSR for victim model), they differ in their robustness against these two watermark removal attacks. For instance, for smooth of GTSRB, after fine-tuning or pruning, 
IDWSR drops under $1\%$, which is below the random guess, however, OoDWSR remains over $67\%$. This phenomenon is also observed for other triggers and datasets. Watermarks injected in OoD samples are much harder to be washed out compared with watermarks injected into i.i.d. samples. 
Due to different distributions, 
fine-tuning or pruning will have a smaller impact on OoD samples compared with i.i.d. samples.

\begin{table}
\vspace{-0.1in}
    \centering
    \small
    \begin{tabular}{p{1.4cm}c|ccc|ccc}
    \toprule
    Trigger&Training &\multicolumn{3}{c}{Victim model}&\multicolumn{3}{c}{Suspect model}\\
    &data&Acc&IDWSR&OoDWSR&Acc&IDWSR&OoDWSR\\
    \hline
    \multirowcell{3}{trojan\_wm}&clean&0.9400&0.0639&0.0487&0.8646&0.0864&0.0741\\
    &ID&0.9378&1.0000&0.9997&0.8593&0.0413&0.0195\\
    &OoD&0.9102&0.9768&0.9566&0.8706&0.4434&\textbf{0.5752}\\
    \hline
    \multirowcell{3}{trojan\_8x8}&clean&0.9400&0.0161&0.0481&0.8646&0.0323&0.0610\\
    &ID&0.9393&0.9963&0.9992&0.8598&0.0342&0.0625\\
    &OoD&0.9178&0.9328&0.9423&0.8675&0.0782&\textbf{0.2419}\\
    \bottomrule
    \end{tabular}
    \caption{Comparison of watermarking methods against fine-tuning watermark removal using different training data.
    OoD injection is much more robust compared with i.i.d. injection.
    }
    \label{tab:id_ood}
    \vspace{-0.25in}
\end{table}

To further verify our intuition, we also compare our method (OoD) with traditional backdoor-based methods using i.i.d. data (ID) for data poisoning on CIFAR-10.
We use RT-AL which is the strongest attack in \cref{tab:finetune} as an example. The results are shown
in \cref{tab:id_ood}. Note that ID poison and the proposed OoD poison
adopt IDWSR and OoDWSR 
as the success rate for the injection watermark, respectively. Clean refers to the pre-trained model without watermark injection.
With only one single OoD image for watermark injection, we can achieve comparable results as ID poisoning which utilizes the entire ID training set. After RT-AL, the watermark success rate drops to $4.13\%$ and $3.42\%$, respectively for ID poison, while drops to $57.52\%$ and $24.19\%$ for OoD poison, which verifies that our proposed method is also much more robust against watermark removal attacks.

\begin{table}[htbp!]
    \centering
    \small
    \scalebox{0.9}{
    \begin{tabular}{cccccccccc}
    \toprule
    Dataset&Trigger&\multicolumn{3}{c}{Victim model}&\multicolumn{3}{c}{Suspect model}&p-value\\
    &&Acc&IDWSR&OoDWSR&Acc&IDWSR&OoDWSR\\
    \hline
    \multirowcell{2}{CIFAR-10}&{trojan\_wm}&{0.9102}&{0.9768}&{0.9566}&0.8485&0.9684&0.9547&0.0000 \\
    &{trojan\_8x8}&{0.9178}&{0.9328}&{0.9423}&0.8529&0.8882&0.9051&0.0000\\

    \hline
    \multirowcell{2}{CIFAR-100}&{trojan\_8x8}&{0.6978}&{0.7024}&{0.8761}&0.5309&0.5977&0.7040&0.0059\\
    &{l0\_inv}&{0.6948}&{0.7046}&{0.5834}&0.5200&0.0162&0.0622&0.0019  \\
    \hline
     \multirowcell{2}{GTSRB}&{smooth}&{0.9146}&{0.1329}&{0.9442}&0.6575&0.1386&0.9419&7.5891e-11\\
    &{trojan\_wm}&{0.9089}&{0.7435}&{0.7513}&0.6379&0.7298&0.7666&2.6070e-21\\

    \bottomrule
    \end{tabular}
    }
    \caption{Evaluation of watermarking against model extraction watermark removal on three datasets. 
    }
    \label{tab:extraction_extended}
    \vspace{-0.25in}
\end{table}

\subsection{Defending Against Model Extraction}
We evaluate the robustness of our proposed method against model extraction  in \cref{tab:extraction_extended}.
By conducting model extraction, the standard accuracy drops $6\%$ on the model pre-trained on CIFAR-10, and drops 
more than $10\%$ on the other two datasets.  Re-training from scratch makes it hard for the suspect model to resume the original model's utility using an OoD dataset and soft labels querying from the watermarked model. OoDWSR is still over $90\%$ and $76\%$ for CIFAR-10 and GTSRB, respectively. Although OoDWSR is $6.22\%$ for l0\_inv, it is still well above $0.02\%$, which is observed for the non-watermarked model. All the datasets also have a p-value close to $0$. All the above observations indicate that the re-training-based extracted model has a high probability of being a copy of our model. One possible reason for these re-training models still extracting the watermark is that during re-training, the backdoor information hidden in the soft label queried by the IP infringers can also embed the watermark in the extracted model. The extracted model will behave more similarly to the victim model as its decision boundary gradually approaches that of the victim model.

\begin{figure}
    \centering
    \vspace{-0.1in}
    \begin{minipage}{0.48\textwidth}
    \begin{subfigure}{0.48\textwidth}
            \centerline{\includegraphics[width=3.2cm]{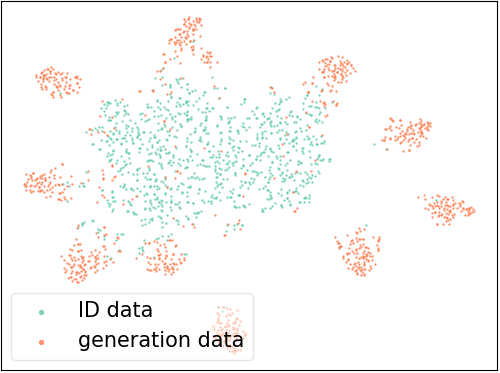}}
        \subcaption{\textbf{Before} fine-tuning.}
        \end{subfigure}  \hfil
        \begin{subfigure}{0.48\textwidth}
            \centerline{\includegraphics[width=3.2cm]{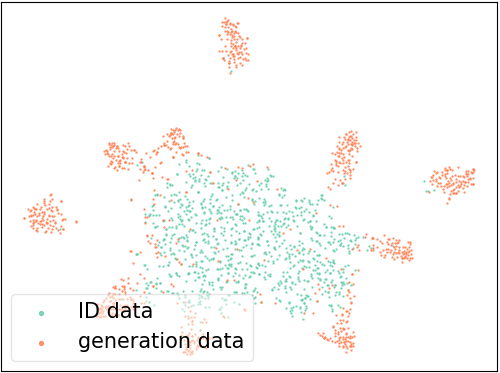}}
        \subcaption{\textbf{After} fine-tuning.}
        \end{subfigure}  \hfil
    \caption{The distribution of OoD and ID samples. 
    Generation data denotes augmented OoD samples from a single OoD image.}
    \label{fig:distribution}
    \end{minipage}
    \hfill
    \begin{minipage}{0.48\textwidth}
        \begin{subfigure}{0.46\textwidth}
            \centerline{\includegraphics[width=4cm]{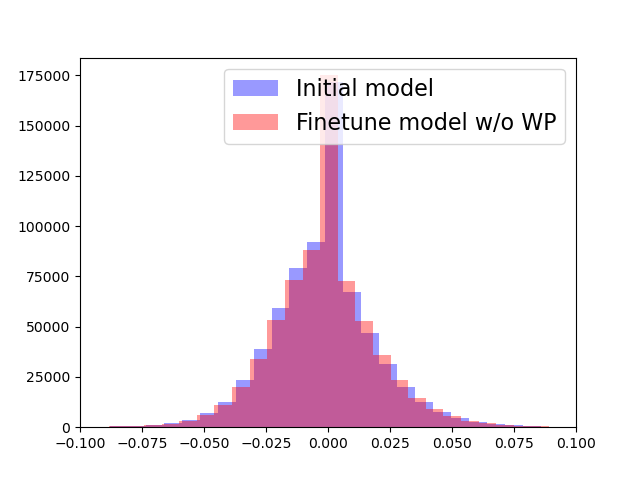}}
            \vspace{-0.1in}
            \subcaption{Without WP.}
        \end{subfigure} 
        \hfill
        \begin{subfigure}{0.46\textwidth}
            \centerline{\includegraphics[width=4cm]{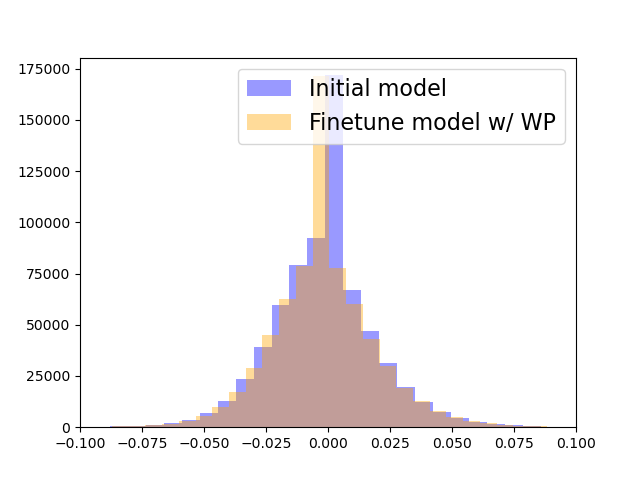}}
            \vspace{-0.1in}
            \subcaption{With WP.}
        \end{subfigure} 
    \caption{Weight distribution for model w/ and w/o WP. The x-axis is the parameter values, and the y-axis is the number of parameters.}
    \label{fig:weight}
    \end{minipage}
    \vspace{-0.2in}
\end{figure}
\subsection{Qualitative Studies}
\textbf{Distribution of generated OoD samples and ID samples.}
We first augment an unlabeled OoD dataset, and then assign predicted labels to them using the model pre-trained on clean CIFAR-10 data. According to the distribution of OoD and ID samples before and after our watermark fine-tuning as shown in \cref{fig:distribution}, we can observe that the OoD data drawn from one image lies close to ID data with a small gap. After a few epochs of fine-tuning, some of the OoD data is drawn closer to ID, but still maintains no overlap. This can help us successfully implant watermarks to the pre-trained model while maintaining the difference between ID and OoD data. In this way, when our model is fine-tuned with clean ID data by attackers, the WSR on the OoD data will not be easily erased.

\textbf{Effects of different OoD images for watermark injection.}
In \cref{tab:ood_image}, we use different source images to generate surrogate datasets and inject watermarks into a pre-trained model. 
The model is pre-trained on CIFAR-10. From these results, we observe that the choice of the OoD image for injection is also important. 
Dense images such as ``City" and ``Animals" can produce higher OoDWSR than the sparse image ``Bridge", since more knowledge is included in the visual representations of dense source images. Thus, dense images perform better for backdoor-based watermark injection. This observation is also consistent with some 
previous arts~\citep{asano2022extrapolating,asano2019critical} about single image representations, which found that dense images perform better for model distillation or self-supervised learning.

\begin{table}[]
    \centering
    \small
     \begin{minipage}{0.43\textwidth} 
    \setlength\tabcolsep{0.55mm}
    \scalebox{0.85}
    {
    \begin{tabular}{ccccc}
    \toprule
    OoD Image&Trigger&Acc&IDWSR&OoDWSR\\
    \hline
    \multirowcell{2}{City}&trojan\_wm&0.9102&0.9768&0.9566  \\
    &trojan\_8x8&0.9178&0.9328&0.9423\\
    \hline
    \multirowcell{2}{Animals}&trojan\_wm&0.9072&0.9873&\textbf{0.9880}\\
    &trojan\_8x8&0.9176&0.9251&0.9622\\
    \hline
    \multirowcell{2}{Bridge}&trojan\_wm&0.9207&0.8749&0.7148\\
    &trojan\_8x8&0.9172&0.7144&0.7147\\
    \bottomrule
    \end{tabular}
    }
    \caption{Watermark injection using different OoD images. %
    }
    \label{tab:ood_image}
    \end{minipage}
    \hfill
    \begin{minipage}{0.54\textwidth}
    \setlength{\tabcolsep}{0.45mm}{
     \scalebox{0.85}
    {
    \begin{tabular}{p{1.4cm}ccccccc}
    
    \toprule
    Trigger&WP&\multicolumn{3}{c}{Victim model}&\multicolumn{3}{c}{Suspect model}\\
    &&Acc&IDWSR&OoDWSR&Acc&IDWSR&OoDWSR\\
    \hline
    \multirowcell{2}{trojan\_wm}&w/o&0.9264&0.9401&0.9490&0.8673&0.1237&0.1994\\
    &w/&0.9102&0.9768&0.9566&0.8706&0.4434&\textbf{0.5752}\\
    \hline
    \multirowcell{2}{trojan\_8x8}&w/o&0.9238&0.9263&0.9486&0.8690&0.0497&0.1281\\
    &w/&0.9178&0.9328&0.9423&0.8675&0.0782&\textbf{0.2419}\\
    \bottomrule
    \end{tabular}}}
    \caption{Weight perturbation increases the robustness of the watermarks against removal attacks.}
    \label{tab:wp}
    \end{minipage}
    \vspace{-0.25in}
\end{table}

\textbf{Effects of backdoor weight perturbation.}
We show the results in \cref{fig:weight}. The initial model is WideResNet pre-trained on CIFAR-10, and the fine-tuned model is the model fine-tuning using our proposed method.  If the OoD data is directly utilized to fine-tune the pre-trained models with only a few epochs, the weight distribution is almost identical for pre-trained and fine-tuned models (left figure). According to \cite{garg2020can}, if the parameter perturbations are small, the backdoor-based watermark can be easily removed by fine-tuning or adding random noise to the model’s parameters.  
Our proposed watermark injection WP (right figure) can shift the fine-tuned model parameters from the pre-trained models in a reasonable scale compared with the left one, while still maintaining high standard accuracy and watermark success rate as shown in \cref{tab:wp}. 
Besides, the weight distribution of the perturbed model still follows a normal distribution as the unperturbed model, performing statistical analysis over the model parameters distributions will not be able to erase our watermark.

To show the effects of WP, we conduct the attack RT-AL on CIFAR-10 as an example.
From \cref{tab:wp}, we observe that  WP does not affect the model utility,
and at the same time, it will become more robust against stealing threats, since OoDWSR increases from $19.94\%$ and $12.81\%$ to $57.52\%$ and $24.19\%$, respectively, for two triggers. More results for WP can be referred to \cref{sec:extended_wp}.

\vspace{-0.1in}
\section{Conclusion}
\vspace{-0.1in}

In this paper, we proposed a novel and practical watermark injection method that does not require training data and utilizes a single out-of-distribution image in a sample-efficient and time-efficient manner. 
We designed a robust weight perturbation method to defend against watermark removal attacks. 
Our extensive experiments on three benchmarks showed that our method efficiently injected watermarks and was robust against three watermark removal threats. 
Our approach has various real-world applications, such as protecting purchased models by encoding verifiable identity and implanting server-side watermarks in distributed learning when ID data is not available. 

\section*{Acknowledgement}

This material is based in part upon work supported by the
National Science Foundation under Grant IIS-2212174, IIS-1749940, Office of Naval Research N00014-20-1-2382, N00014-24-1-2168, and National Institute on Aging (NIA) RF1AG072449.
The work of Z. Wang is in part supported by the National Science Foundation under Grant IIS2212176.

\bibliography{auto_gen}
\bibliographystyle{iclr2024_conference}

\appendix
\clearpage
\section{Methodology supplementaries}
\subsection{Extended watermark injection results}\label{sec:extended_injection}
\begin{figure*}[htbp]  
    \begin{center}
        \begin{subfigure}{0.33\textwidth}
            \centerline{\includegraphics[width=6cm]{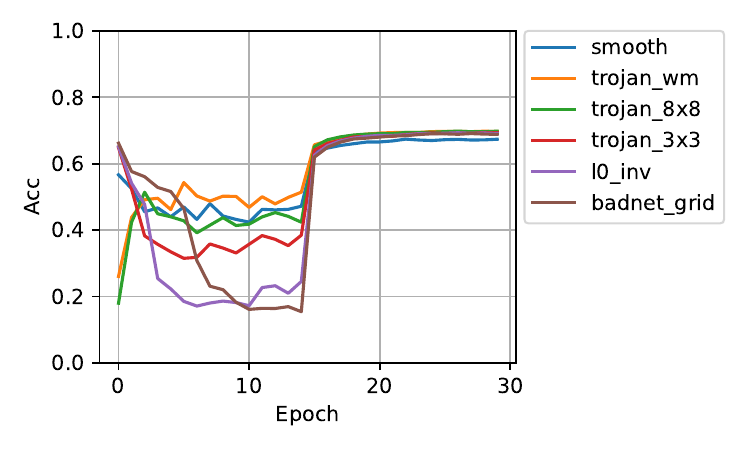}}
        \subcaption{CIFAR-100 Acc.}
        \end{subfigure}  \hfil
       \begin{subfigure}{0.33\textwidth}
            \centerline{\includegraphics[width=6cm]{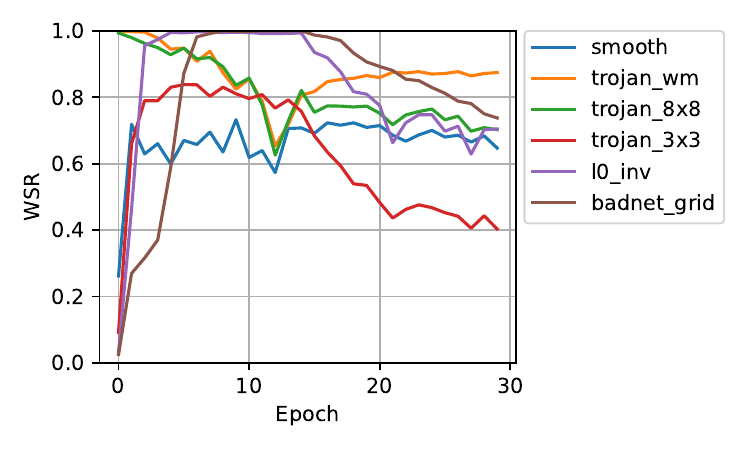}}
        \subcaption{CIFAR-100 ID WSR.}
        \end{subfigure}  \hfil
        \begin{subfigure}{0.33\textwidth}
            \centerline{\includegraphics[width=6 cm]{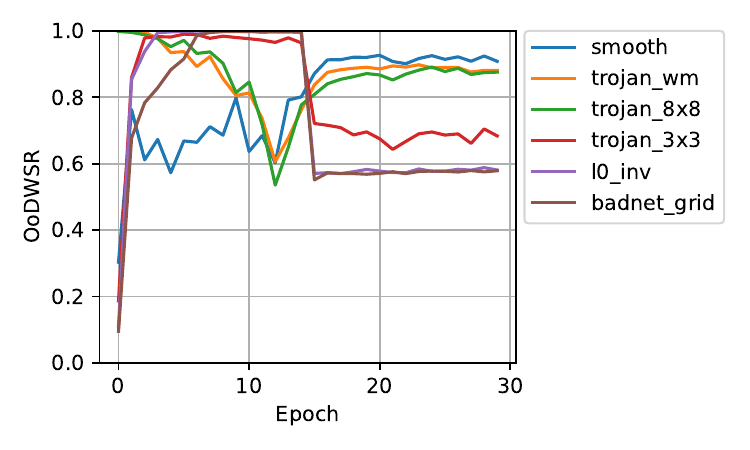}}
        \subcaption{CIFAR-100 OoD WSR.}
        \end{subfigure}  \hfil
        \begin{subfigure}{0.33\textwidth}
            \centerline{\includegraphics[width=6cm]{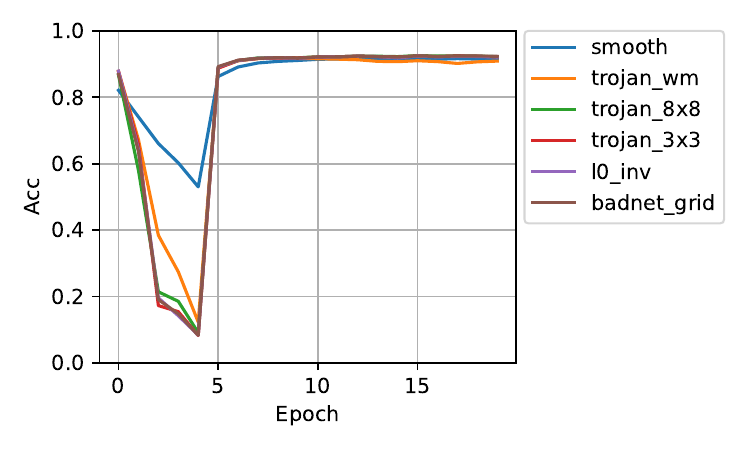}}
        \subcaption{GTSRB Acc.}
        \end{subfigure}  \hfil
       \begin{subfigure}{0.33\textwidth}
            \centerline{\includegraphics[width=6cm]{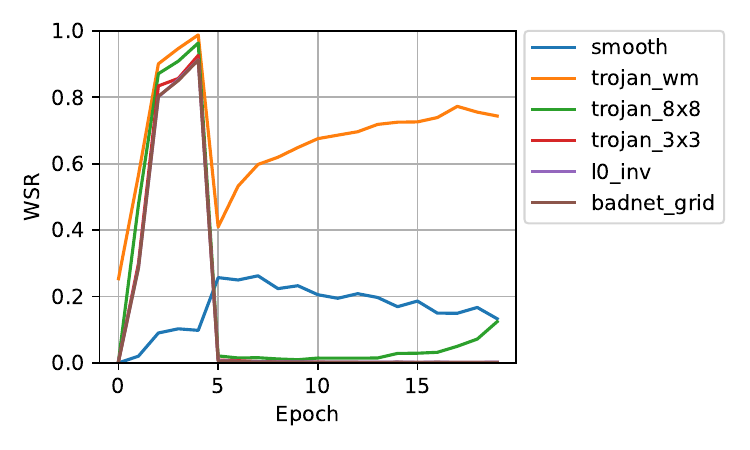}}
        \subcaption{GTSRB ID WSR.}
        \end{subfigure}  \hfil
        \begin{subfigure}{0.33\textwidth}
            \centerline{\includegraphics[width=6 cm]{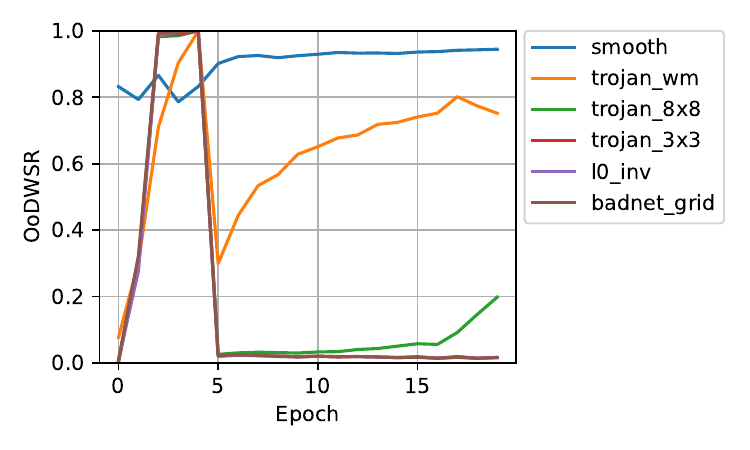}}
        \subcaption{GTSRB OoD WSR.}
        \end{subfigure} 
    \caption{Acc, ID WSR, and OoD WSR for watermark injection. The watermarks are injected quickly with high accuracy and OoDWSR. Triggers with the highest OoDWSR and accuracy degradation of less than $3\%$ are selected for each dataset.
    }\label{fig:inject_addition}
    \end{center}
        \vspace{-0.1in}
\end{figure*}

\textbf{Additional watermark injection results.}
\cref{fig:inject_addition} shows the watermark injection for CIFAR-100 and GTSRB. It only takes $20$ epochs for CIFAR-100 and GTSRB 
to achieve stable high standard accuracy and OoDWSR. 
The highest OoDWSR for CIFAR-100, and GTSRB are $0.8761$, and $0.9442$, respectively, with standard accuracy degradation of less than $3\%$.

\subsection{Extended weight perturbation results}\label{sec:extended_wp}

\textbf{Additional experiments about the effects of WP.}
\cref{tab:wp_ftal} shows the results for fine-tuning method FT-AL. We observe that by applying WP, after fine-tuning, we can increase OoDWSR from $0.7305$ and $0.8184$ to $0.9678$ and $0.9797$, respectively. 
\begin{table}[htbp!]
    \centering
    \small
    \setlength\tabcolsep{2.2 pt}
    \begin{tabular}{p{1.4cm}ccccccc}
    
    \toprule
    Trigger&WP&\multicolumn{3}{c}{Victim model}&\multicolumn{3}{c}{Suspect model}\\
    &&Acc&IDWSR&OoDWSR&Acc&IDWSR&OoDWSR\\
    \hline
    \multirowcell{2}{trojan\_wm}&w/o&0.9264& 0.9401 &0.9490&0.9226&0.6327&0.7305\\
    &w/&0.9102&0.9768&0.9566&0.9191&0.9769&\textbf{0.9678}\\
    \hline
    \multirowcell{2}{trojan\_8x8}&w/o&0.9238& 0.9263 &0.9486&0.9223&0.5304&0.8184\\
    &w/&0.9178&0.9328&0.9423&0.9187&0.9533&\textbf{0.9797}\\
    \bottomrule
    \end{tabular}
    \caption{Weight perturbation increases the robustness of the watermarks against removal attacks (FT-AL).}
    \label{tab:wp_ftal}
    \vspace{-0.2in}
\end{table}

\begin{figure}[htbp]
    \centering
    \begin{subfigure}{0.33\textwidth}
            \centerline{\includegraphics[width=4.5cm]{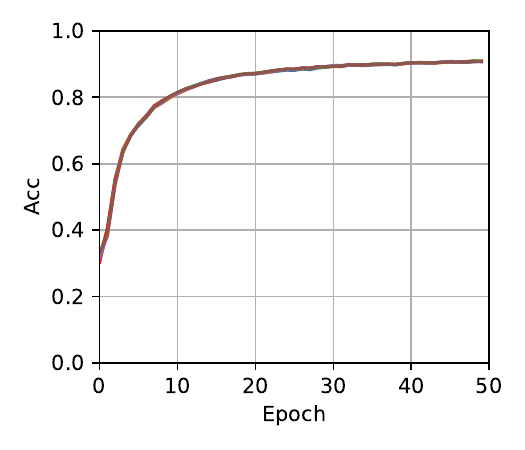}}
        \subcaption{Acc for model w/o WP.}
        \end{subfigure} \hfil
    \begin{subfigure}{0.33\textwidth}
            \centerline{\includegraphics[width=6.5cm]{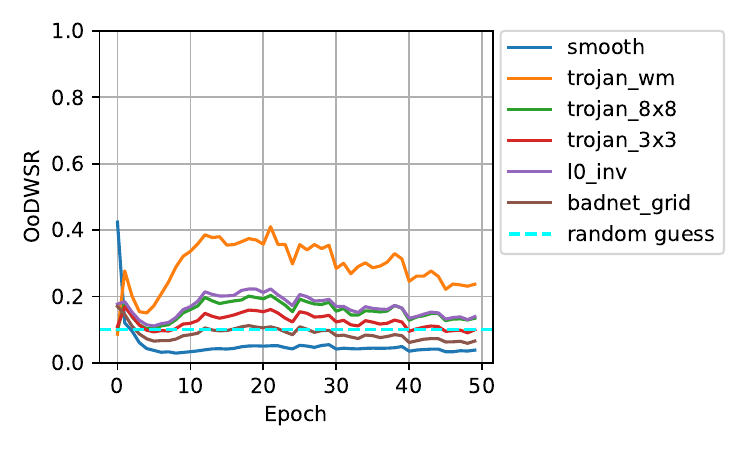}}
        \subcaption{OoDWSR for model w/o WP.}
        \end{subfigure} \hfil
        \begin{subfigure}{0.33\textwidth}
            \centerline{\includegraphics[width=4.5cm]{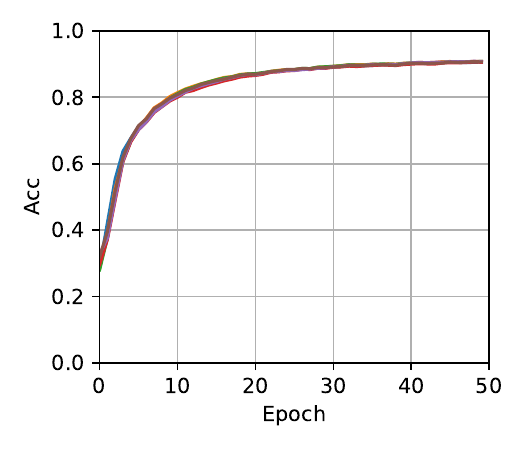}}
        \subcaption{Acc for model w/ WP.}
        \end{subfigure} \hfil
        \begin{subfigure}{0.33\textwidth}
            \centerline{\includegraphics[width=6.5cm]{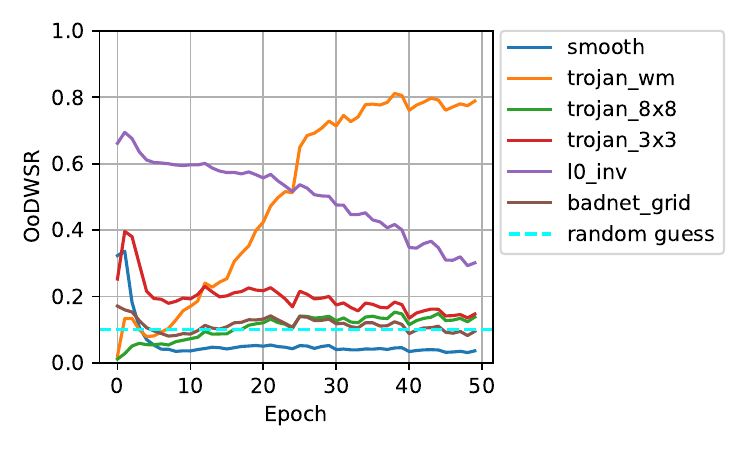}}
        \subcaption{OoDWSR for model w/ WP.}
        \end{subfigure} 
    \caption{Effects of weight perturbation against RT-AL with $100\%$ training data on CIFAR-10.}
    \label{fig:wp}
\end{figure}

To further verify the robustness of our proposed weight perturbation, in \cref{fig:wp}, we also show the results of a much more challenging setting, i.e., RT-AL with $100\%$ training data. With more data for fine-tuning, RT-AL can obtain an average standard accuracy of $0.9069$ and $0.9074$, respectively for the model w/ and w/o WP. With comparable standard accuracy, WP can increase OoDWSR by $55.45\%$, $0.35\%$, $5.02\%$, $16.23\%$ for trojan\_wm, trojan\_8x8, trojan\_3x3, l0\_inv, respectively. With WP, during the fine-tuning process, OoDWSR will remain more stable or even increase with the increase of standard accuracy. These results demonstrate that the proposed WP can help our injected watermark be more robust and persistent even under more challenging stealing threats.

\begin{figure}[htbp!]
    \centering
    \begin{subfigure}{0.23\textwidth}
            \centerline{\includegraphics[width=4.5cm]{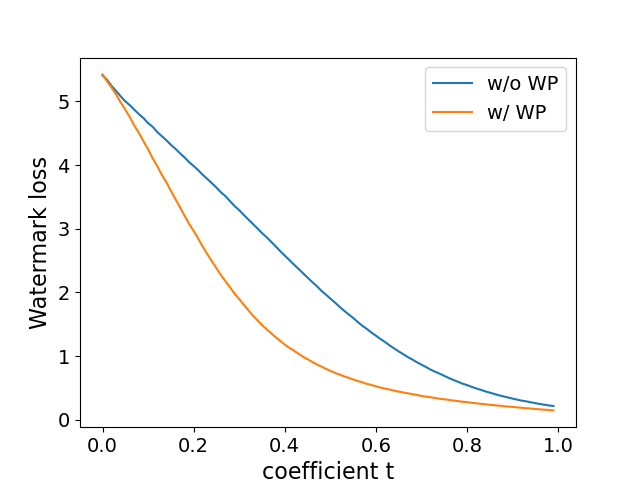}}
        \subcaption{trojan\_wm.}
        \end{subfigure} \hfil
    \begin{subfigure}{0.23\textwidth}
            \centerline{\includegraphics[width=4.5cm]{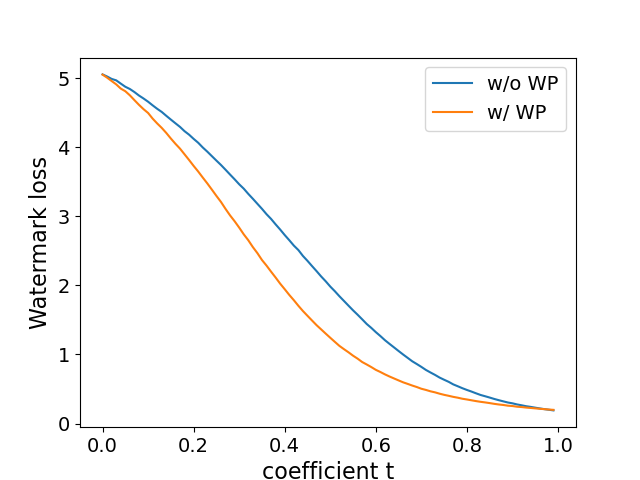}}
        \subcaption{trojan\_8x8.}
        \end{subfigure} \hfil
       
    \caption{Watermark loss landscape from pre-trained model to watermarked model on CIFAR-10.}
    \label{fig:landscape}
\end{figure}

\textbf{Watermark loss landscape for WP.}
To get the watermark loss landscape around the watermarked model $\cM_w(\theta_w)$, we interpolate between the initial pre-trained model $\cM(\theta)$ and the watermarked model $\cM_w(\theta_w)$ along the segment $\theta_i = (1-t)\theta + t \theta_w$, where $t\in [0,1]$ with increments of $0.01$. The watermark loss is evaluated on the verification dataset composed of the generated OoD samples. Pre-trained model $\cM(\theta)$ and watermarked model $\cM_w(\theta_w)$ correspond to coefficients of $0$ and $1$, respectively. From \cref{fig:landscape}, we observe that with WP, we can achieve a flatter loss landscape (orange line) around the point of the watermarked model compared with the one without WP (blue line). By maximizing the backdoor loss in Eq. (3) over the perturbation $v$ we can get a flatter landscape of watermark loss around the optimal point of the watermarked model.
For those watermark removal attacks such as fine-tuning or pruning, which might make a minor parameter change to the models, 
a flatter loss landscape could prevent the model from escaping from the watermarked local optimum compared with sharp ones. The loss landscape of trojan\_wm is flatter than trojan\_8x8, thus, the robustness of trojan\_wm is also better than trojan\_8x8 as shown in Table 2 in our paper.

\begin{figure}[htbp!]
    \centering
    \begin{subfigure}{0.24\textwidth}
            \centerline{\includegraphics[width=3.9cm]{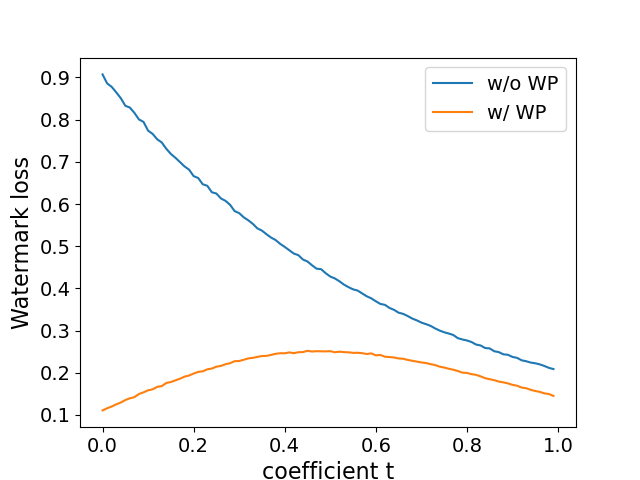}}
        \subcaption{FT-AL for trojan\_wm.}
        \end{subfigure} \hfil
    \begin{subfigure}{0.24\textwidth}
            \centerline{\includegraphics[width=3.9cm]{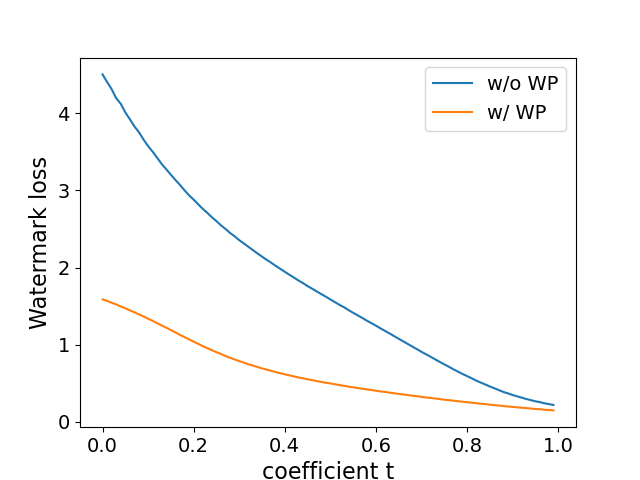}}
        \subcaption{RT-AL for trojan\_wm.}
        \end{subfigure} \hfil
        \begin{subfigure}{0.24\textwidth}
            \centerline{\includegraphics[width=3.9cm]{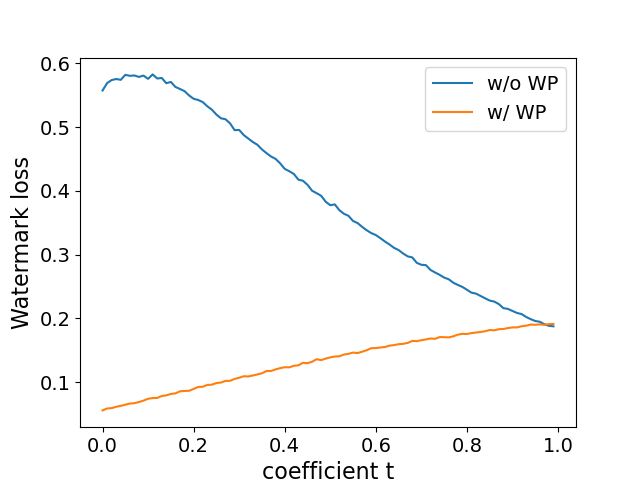}}
        \subcaption{FT-AL for trojan\_8x8.}
        \end{subfigure} \hfil
    \begin{subfigure}{0.24\textwidth}
            \centerline{\includegraphics[width=3.9cm]{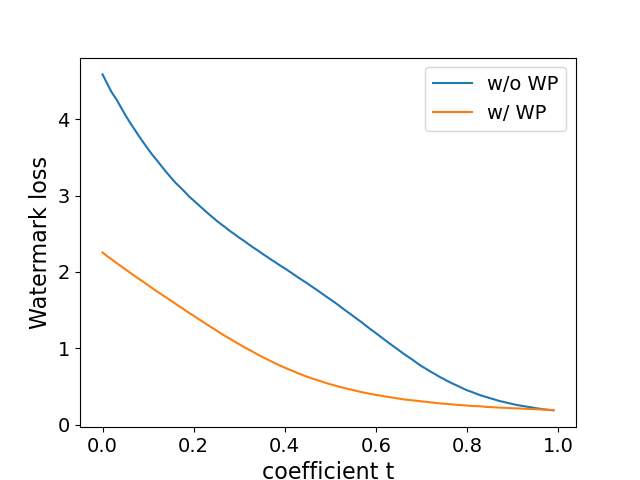}}
        \subcaption{RT-AL for trojan\_8x8.}
        \end{subfigure} \hfil
       
    \caption{Watermark loss landscape from suspect model to watermarked model on CIFAR-10.}
    \label{fig:landscape_finetune}
\end{figure}

To show the relationship between robustness of our proposed method against watermark removal attacks and the loss landscape,  we also interpolate between the suspect model after clean data fine-tuning $\cM_s(\theta_s)$  and the watermarked model $\cM_w(\theta_w)$ along the segment $\theta_i = (1-t)\theta_s + t \theta_w$, where $t\in [0,1]$ with increments of $0.01$. Suspect model $\cM_s(\theta_s)$ and watermarked model $\cM_w(\theta_w)$ correspond to coefficients of $0$ and $1$, respectively. \cref{fig:landscape_finetune} shows the loss for the interpolated model between the suspect model after clean i.i.d. data fine-tuning (FT-AL and RT-AL) and our watermarked model. We observe that the flatness of the loss landscape can lead to a lower watermark loss during fine-tuning, which makes it harder for IP infringers to escape from the local optimum. These results combinedly give an explanation for why our proposed WP can increase the robustness of the watermark against watermark removal attacks.

\subsection{Extended robustness analysis}

\textbf{Defending against OoD detection.} \cite{kim2023margin} found that the adversary may simply reject the query of OoD samples to reject the ownership verification process using the energy-based out-of-distribution detection method~\citep{liu2020energy}. We compute the mean energy score for ID samples and our verification samples, and also report the area under the receiver operating
characteristic curve (AUROC), and the area under the PR curve (AUPR) for ID and OoD classification according to ~\citep{liu2020energy} in \cref{tab:detection}. For the watermarked model pre-trained on CIFAR-100 and GTSRB, the energy scores for ID and verification samples are very close, and the AUROC and AUPR are close to random guesses. For CIFAR-10, despite the differences of energy score, AUPR for OoD detection is only $20\%$, which indicates that it will be very hard for the adversaries to filter out verification samples from ID ones, if verification samples are mixed with other samples while querying the suspect model. Therefore, the query of our verification samples cannot be found out by the IP infringers. A possible reason is that the distribution of our generated verification samples lies closely to the ID ones, since previous work~\citep{asano2022extrapolating} shows that the generated samples from the single OoD image can also be used to train a classifier yielding reasonable performance on the main prediction task.
\begin{table}[htbp!]
    \centering
    \small
    \begin{tabular}{cccccc}
     \toprule
         Dataset&Trigger&Energy of ID samples&Energy of verification samples& AUROC&AUPR\\
         \hline
          \multirowcell{2}{
         CIFAR-10}&trojan\_wm&-5.2880&-14.0207&0.8160&0.2068\\
         &{trojan\_8x8}&-5.1726&-11.2574&0.8107&0.2028\\
         \hline
         \multirowcell{2}
         {CIFAR-100}&{trojan\_8x8}& -14.3086&-13.4729&0.4487&0.0757\\
         &l0\_inv&-13.9304&-11.9693&0.3614&0.0651\\
         \hline
         \multirowcell{2}
         {GTSRB}&smooth&-8.6203&-8.2872&0.4564&0.0715\\
         &trojan\_wm&-8.3042&-8.4407&0.4922&0.0646\\
         \bottomrule
    \end{tabular}
    \caption{Energy-based OoD detection between ID and verification samples. Higher AUROC and AUPR indicate better OoD detection performance.}
    \label{tab:detection}
\end{table}
\subsection{Extended ablation sududies}
\begin{figure}[htbp]
    \begin{center}
    \includegraphics[width=0.5\textwidth]{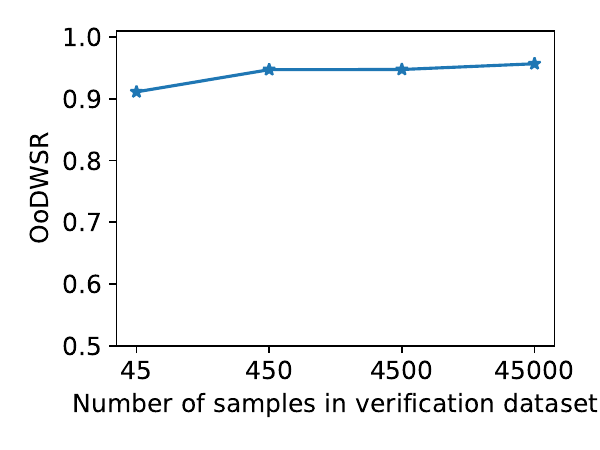}
    \end{center}
     \vspace{-0.2in}
    \caption{OoDWSR 
w.r.t. different numbers of samples in the verification dataset. The model is pre-trained on CIFAR-10, and the trigger pattern is trojan\_wm.}
    \label{fig:oodpercent}
     \vspace{-0.2in}
\end{figure}

\textbf{Effects of different numbers of verification samples.} We generate $45000$ surrogate samples for watermark injection, but we do not need to use all the surrogate data for verification. We report the OoDWSR 
w.r.t. different numbers of samples in the verification dataset in \cref{fig:oodpercent}. According to the figure, only $450$ verification samples are enough for accurate verification.

\textbf{Effects of different number of source OoD images.}
\begin{table}[htbp!]
    \centering
    \small
    \scalebox{1}{
    \begin{tabular}{ccccccccc}
    \toprule
    Source OoD Image&\multicolumn{3}{c}{Victim model}&\multicolumn{3}{c}{Suspect model}&p-value\\
    &Acc&IDWSR&OoDWSR&Acc&IDWSR&OoDWSR\\
    \hline
   {City}&{0.9102}&{0.9768}&{0.9566}&0.9191&0.9769&0.9678&0.0000 \\
  {Animal}&0.9072&0.9873&0.9880&0.9212&0.8309&0.8922&0.0000\\{City+Animal}&0.9059&0.9820&0.9638&0.9209&0.8311&0.8511&0.0000\\

    \bottomrule
    \end{tabular}
    }
    \caption{Effects of different number of source OoD images.
    }
    \label{tab:OoD_number}
    \vspace{-0.1in}
\end{table}
We evaluate the effect of a combination of different source OoD images in \cref{tab:OoD_number}. The model is pre-trained on CIFAR-10, and trojan\_wm is adopted as the trigger pattern. The Suspect model is fine-tuned using FT-AL by the IP infringer. According to the results, for watermark injection (see victim model), with a combination of two OoD images, OoDWSR is between the values for the two combination images. For the robustness against fine-tuning conducted by the IP infringer (see suspect model), OoDWSR even decreases after the combination of the source OoD images. However, one advantage of using multiple OoD images is that it will be harder for the IP infringers to get access to the source OoD image and guess the composition of our validation set, which can further enhance the security of the watermark.

\end{document}